%% file: root.tex
\documentclass[letterpaper, 10 pt, conference]{ieeeconf}  

\IEEEoverridecommandlockouts                              

\usepackage{amsmath} 
\usepackage{amssymb}  
\usepackage[usenames, dvipsnames]{color}
\usepackage[short]{optidef}
\usepackage{mathtools}
\usepackage{algorithm}
\usepackage{algorithmicx}
\usepackage{pgfplots}
\usepackage{graphicx}
\usepackage[abs]{overpic}
\usepackage{booktabs}
\usepackage{caption}
\usepackage{algpseudocode}
\usepackage{hyperref}

\algdef{SE}[DOWHILE]{Do}{doWhile}{\algorithmicdo}[1]{\algorithmicwhile\ #1}%
\usepackage[style=ieee,doi=false,isbn=false,url=false,eprint=false]{biblatex}
\title{\LARGE \bf
ReLU-QP: A GPU-Accelerated Quadratic Programming Solver for Model-Predictive Control
}

\author{Arun L. Bishop$^{1*}$, John Z. Zhang$^{1*}$, Swaminathan Gurumurthy$^1$, Kevin Tracy$^1$, Zachary Manchester$^1$
\thanks{$^1$ Authors are with the Robotics Institute, School of Computer Science, Carnegie Mellon University, Pittsburgh, PA 15213, USA
{\tt\footnotesize \{arunleob, johnzhang, swamig, ktracy, zacm\}@cmu.edu}}%
\thanks{$^*$These authors contributed equally to this paper}
}

\begin{document}

\maketitle
\thispagestyle{empty}
\pagestyle{empty}


\input{sections/abstract}
\input{sections/introduction}
\input{sections/background_related_works}
\input{sections/technical_approach}

\input{sections/MPC}
\input{sections/experiments_and_results}
\input{sections/discussion_and_conclusions}


\printbibliography

\end{document}

%% file: sections/abstract.tex
\begin{abstract}

We present ReLU-QP, a GPU-accelerated solver for quadratic programs (QPs) that is capable of solving high-dimensional control problems at real-time rates. ReLU-QP is derived by exactly reformulating the Alternating Direction Method of Multipliers (ADMM) algorithm for solving QPs as a deep, weight-tied neural network with rectified linear unit (ReLU) activations. This reformulation enables the deployment of ReLU-QP on GPUs using standard machine-learning toolboxes. 
We evaluate the performance of ReLU-QP across three model-predictive control (MPC) benchmarks: stabilizing random linear dynamical systems with control limits, balancing an Atlas humanoid robot on a single foot, and tracking whole-body reference trajectories on a quadruped equipped with a six-degree-of-freedom arm.
These benchmarks indicate that ReLU-QP is competitive with state-of-the-art CPU-based solvers for small-to-medium-scale problems and offers order-of-magnitude speed improvements for larger-scale problems.

\end{abstract}

%% file: sections/introduction.tex
\section{Introduction}

Convex quadratic programming is a core technology in many areas of robotic control, including inverse kinematics \cite{shankar_quadratic_2015}, contact dynamics \cite{redon_gauss_2002}, actuator design \cite{spielberg_functional_2017}, and model-predictive control (MPC) \cite{kouvaritakis_model_2016}. In many cases, problem instances with thousands or tens of thousands of variables need to be solved in milliseconds to achieve real-time performance. 

In this paper, we focus on MPC, which places particularly challenging demands on solvers by requiring large problems to be solved robustly and very quickly. Model-predictive control (MPC) is a widely used control strategy that solves a receding-horizon optimization problem while reasoning about linear (or linearized) system dynamics and additional linear equality and inequality constraints. Successful applications of MPC include autonomous driving \cite{falcone_predictive_2007}, rocket landing \cite{acikmese_lossless_2013}, and legged locomotion \cite{di_carlo_dynamic_2018}. MPC problems frequently include hundreds to tens of thousands of variables, with solver time complexity scaling quadratically with state and control dimensions and linearly with the time horizon.

To reduce solve times, it is common practice to use simplified point-mass \cite{acikmese_convex_2007, acikmese_lossless_2013} or centroidal \cite{bledt_mit_2018, kuindersma_optimization-based_2016} models that limit the controller's ability to reason about system dynamics and state and control constraints, such as joint and torque limits. When whole-body models are used in MPC controllers, the increased computational burden results in longer solve times and lower sample rates, requiring a faster low-level PD controller to compensate \cite{neunert_whole-body_2018}, and often producing more conservative and less robust performance.

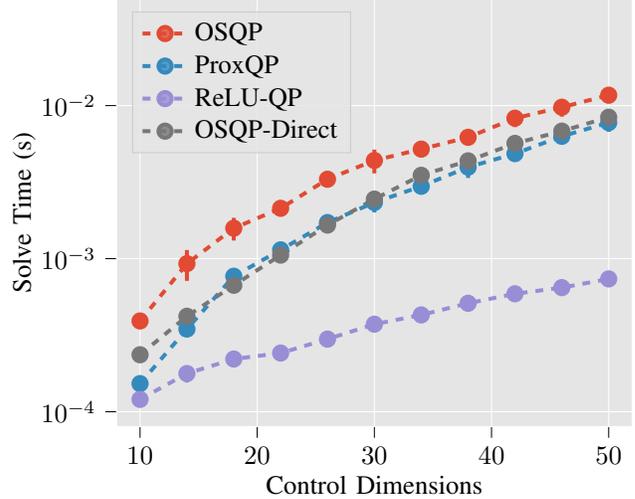
\begin{figure}
    \centering
    \input{figures/sweep_state_control_dims}
    \caption{MPC solve times for random linear systems with control limits (horizon = $40$, $3$:$1$ state-to-control dimension ratio). ReLU-QP is an order of magnitude faster on high-dimensional control problems. The preconditioned condensed formulation is used except \emph{OSQP-Direct}, which solves the sparse direct formulation.}
    \label{fig:random_mpc}
\end{figure}

Current methods for solving quadratic programs generally fall into three categories: interior-point methods, active-set methods, and the closely related augmented-Lagrangian method and alternating-direction method of multipliers (ADMM). Interior-point solvers like GUROBI \cite{gurobi}, MOSEK \cite{aps_mosek_nodate}, and IPOPT \cite{wachter_implementation_2006}, offer robust convergence with a few iterations of Newton's method, but each iteration requires the expensive solution of a linear system. Interior-point methods are also difficult to warm start, making them less well suited to MPC applications where similar problem instances are encountered at each time step.

Active-set methods like QuadProg \cite{goldfarb_numerically_1983} and qpOASES \cite{ferreau_qpoases_2014} treat active inequality constraints as equalities and remove inactive ones from the problem. They are easy to warm start and can be very fast if the active set is guessed correctly, but their worst-case time complexity is combinatorial in the number of constraints \cite{NoceWrig06}

Augmented-Lagrangian and ADMM methods, including ProxQP \cite{bambade_prox-qp_2022}, OSQP \cite{stellato_osqp_2020}, and ALTRO \cite{howell_altro_2019}, have recently gained popularity in the robotics community for MPC. These solvers are easy to warm start and converge quickly to coarse tolerances, but generally have slow ``tail convergence'' when trying to achieve tight constraint tolerances.

While most existing MPC algorithms run on CPUs, there has been significant recent interest in exploiting the parallel computing capabilities of GPUs to speed up MPC. One such approach is model predictive path integral control (MPPI) \cite{williams_information_2017, pravitra_1-adaptive_2020}, which randomly samples control sequences and performs thousands of simulation rollouts in parallel to determine the next control action. While fast, MPPI can struggle to stabilize open-loop unstable systems and suffers from poor sample efficiency, making it difficult to scale to high-dimensional systems.

There have also been recent efforts to parallelize ADMM algorithms. A GPU version of OSQP uses an indirect conjugate-gradient method in place of a direct matrix factorization to exploit fast parallel matrix-vector products \cite{schubiger_gpu_2020}. Another parallelized ADMM implementation pre-computes matrix inverses and relies on custom CUDA kernels for computing matrix-vector products on the GPU, but otherwise maintains the conventional sequential structure of ADMM \cite{yu_efficient_2017}.

In this work, we introduce ReLU-QP, a GPU-accelerated quadratic programming solver that achieves speed-ups of an order of magnitude over existing solvers on large-scale problems. ReLU-QP achieves this by analytically mapping the sequential updates of the ADMM algorithm into the structure of a deep, weight-tied neural network with rectified linear unit (ReLU) activations. With this mapping, each iteration of the ADMM algorithm corresponds to a single layer of the network, comprising a dense matrix-vector product followed by a projection onto the positive orthant (clamping negative values to zero). 

ReLU-QP can handle large MPC problem instances, enabling the use of detailed whole-body models subject to joint and torque limits, friction cones, and other complex constraints.

In spite of a large number of constraints and high-dimensional state and input spaces in these problems, ReLU-QP is able to achieve kilohertz solution rates, allowing us to replace the standard hierarchical control stack --- in which an MPC controller is cascaded with low-level PD controllers --- with a single, unified, MPC controller that can more fully reason about a robot's dynamics for better performance, safety, and robustness. The simplicity of ReLU-QP also makes it easy to implement and deploy with standard machine-learning toolboxes such as PyTorch, TensorFlow, and JAX. 

Our contributions are: 
\begin{itemize}
    \item An analytically exact transformation of the ADMM algorithm into a deep, weight-tied, neural network
    \item Open-source implementations of the ReLU-QP solver in PyTorch, JAX, and Julia
    \item Experimental validation of ReLU-QP on several representative model-predictive control benchmarks with comparisons to state-of-the-art CPU-based solvers
\end{itemize}

Our paper is organized as follows: In Section \ref{sec:background}, we review quadratic programs, the ADMM algorithm, and MPC. In Section \ref{sec:reluqp}, we derive the mapping between ADMM and a deep, weight-tied neural network and present the ReLU-QP solver. In Section \ref{sec:mpc}, we discuss the details of our MPC implementation for use with ReLU-QP. In Section \ref{sec:experiments} we benchmark ReLU-QP against OSQP and ProxQP on dense QPs and three MPC applications: random linear problems, the Atlas humanoid balancing on one foot, and a quadruped equipped with a 6-DoF arm tracking whole-body reaching motions. Finally, we summarize our results, discuss the limitations of our work, and outline future research directions in Section \ref{sec:conclusions}.

%% file: figures/sweep_state_control_dims.tex
\begin{tikzpicture}

\definecolor{chocolate2267451}{RGB}{226,74,51}
\definecolor{gainsboro229}{RGB}{229,229,229}
\definecolor{gray119}{RGB}{119,119,119}
\definecolor{lightgray204}{RGB}{204,204,204}
\definecolor{mediumpurple152142213}{RGB}{152,142,213}
\definecolor{steelblue52138189}{RGB}{52,138,189}

\begin{axis}[
axis background/.style={fill=gainsboro229},
axis line style={white},
legend cell align={left},
legend style={
  fill opacity=0.8,
  draw opacity=1,
  text opacity=1,
  at={(0.03,0.97)},
  anchor=north west,
  draw=lightgray204,
  fill=gainsboro229
},
log basis y={10},
tick align=outside,
tick pos=left,
x grid style={white},
xlabel={Control Dimensions},
xmajorgrids,
xmin=8, xmax=52,
xtick style={color=black},
y grid style={white},
ylabel={Solve Time (s)},
ymajorgrids,
ymin=8e-05, ymax=0.05,
ymode=log,
ytick style={color=black},
ytick={1e-4, 1e-3, 1e-2}
]
\path [draw=chocolate2267451, line width=1.5pt]
(axis cs:10,0.000359192459253281)
--(axis cs:10,0.000424374528501821);

\path [draw=chocolate2267451, line width=1.5pt]
(axis cs:14,0.000716391784471994)
--(axis cs:14,0.00113928770940556);

\path [draw=chocolate2267451, line width=1.5pt]
(axis cs:18,0.00131374025748583)
--(axis cs:18,0.00184827673026927);

\path [draw=chocolate2267451, line width=1.5pt]
(axis cs:22,0.00200208754520016)
--(axis cs:22,0.00226402524663657);

\path [draw=chocolate2267451, line width=1.5pt]
(axis cs:26,0.00297165275124772)
--(axis cs:26,0.00363482589365024);

\path [draw=chocolate2267451, line width=1.5pt]
(axis cs:30,0.00360705437290561)
--(axis cs:30,0.00515893961893112);

\path [draw=chocolate2267451, line width=1.5pt]
(axis cs:34,0.00474887846806566)
--(axis cs:34,0.00564325802173026);

\path [draw=chocolate2267451, line width=1.5pt]
(axis cs:38,0.00547324066233758)
--(axis cs:38,0.00694140340296854);

\path [draw=chocolate2267451, line width=1.5pt]
(axis cs:42,0.0072363363349999)
--(axis cs:42,0.00930942635887765);

\path [draw=chocolate2267451, line width=1.5pt]
(axis cs:46,0.00847033442222734)
--(axis cs:46,0.0110424452349155);

\path [draw=chocolate2267451, line width=1.5pt]
(axis cs:50,0.0103513693249379)
--(axis cs:50,0.0130409454913886);

\path [draw=steelblue52138189, line width=1.5pt]
(axis cs:10,0.000132887143214706)
--(axis cs:10,0.000172491354744478);

\path [draw=steelblue52138189, line width=1.5pt]
(axis cs:14,0.000317942483028858)
--(axis cs:14,0.000375949174113999);

\path [draw=steelblue52138189, line width=1.5pt]
(axis cs:18,0.000727811161532013)
--(axis cs:18,0.000807626634386354);

\path [draw=steelblue52138189, line width=1.5pt]
(axis cs:22,0.000996130016575517)
--(axis cs:22,0.00128629771403673);

\path [draw=steelblue52138189, line width=1.5pt]
(axis cs:26,0.00162423801600484)
--(axis cs:26,0.00181898852277067);

\path [draw=steelblue52138189, line width=1.5pt]
(axis cs:30,0.00200465741562277)
--(axis cs:30,0.00265099367825478);

\path [draw=steelblue52138189, line width=1.5pt]
(axis cs:34,0.00265937754512388)
--(axis cs:34,0.00328172371201898);

\path [draw=steelblue52138189, line width=1.5pt]
(axis cs:38,0.00336754039785829)
--(axis cs:38,0.00455353713683559);

\path [draw=steelblue52138189, line width=1.5pt]
(axis cs:42,0.00435370915886988)
--(axis cs:42,0.00535631998398726);

\path [draw=steelblue52138189, line width=1.5pt]
(axis cs:46,0.00578036496508346)
--(axis cs:46,0.00681936425940634);

\path [draw=steelblue52138189, line width=1.5pt]
(axis cs:50,0.00674423381043123)
--(axis cs:50,0.00873150409160958);

\path [draw=mediumpurple152142213, line width=1.5pt]
(axis cs:10,0.000119679519810237)
--(axis cs:10,0.000121104153771401);

\path [draw=mediumpurple152142213, line width=1.5pt]
(axis cs:14,0.000155022830570442)
--(axis cs:14,0.000200166722729971);

\path [draw=mediumpurple152142213, line width=1.5pt]
(axis cs:18,0.000214115603022753)
--(axis cs:18,0.000227786895715631);

\path [draw=mediumpurple152142213, line width=1.5pt]
(axis cs:22,0.000229212179870512)
--(axis cs:22,0.000254179558664222);

\path [draw=mediumpurple152142213, line width=1.5pt]
(axis cs:26,0.000288729966903756)
--(axis cs:26,0.000307515716288956);

\path [draw=mediumpurple152142213, line width=1.5pt]
(axis cs:30,0.000366868930138803)
--(axis cs:30,0.000379256652293362);

\path [draw=mediumpurple152142213, line width=1.5pt]
(axis cs:34,0.00040116739972957)
--(axis cs:34,0.00045683619128667);

\path [draw=mediumpurple152142213, line width=1.5pt]
(axis cs:38,0.00048393628533137)
--(axis cs:38,0.000538258913707464);

\path [draw=mediumpurple152142213, line width=1.5pt]
(axis cs:42,0.000567151483439694)
--(axis cs:42,0.000612032058704278);

\path [draw=mediumpurple152142213, line width=1.5pt]
(axis cs:46,0.000622135681051993)
--(axis cs:46,0.000671534981332183);

\path [draw=mediumpurple152142213, line width=1.5pt]
(axis cs:50,0.000730253698266865)
--(axis cs:50,0.00074501970766187);

\path [draw=gray119, line width=1.5pt]
(axis cs:10,0.000231058226748535)
--(axis cs:10,0.000240246381251465);

\path [draw=gray119, line width=1.5pt]
(axis cs:14,0.000415239867397219)
--(axis cs:14,0.000425625532602781);

\path [draw=gray119, line width=1.5pt]
(axis cs:18,0.000665426877774573)
--(axis cs:18,0.000674824870225427);

\path [draw=gray119, line width=1.5pt]
(axis cs:22,0.00103424116607779)
--(axis cs:22,0.00107559070192221);

\path [draw=gray119, line width=1.5pt]
(axis cs:26,0.00158342623470614)
--(axis cs:26,0.00173131184529386);

\path [draw=gray119, line width=1.5pt]
(axis cs:30,0.00241203831987717)
--(axis cs:30,0.00248679180012283);

\path [draw=gray119, line width=1.5pt]
(axis cs:34,0.00343531161344212)
--(axis cs:34,0.00357526996655788);

\path [draw=gray119, line width=1.5pt]
(axis cs:38,0.0043062495531484)
--(axis cs:38,0.0044049283668516);

\path [draw=gray119, line width=1.5pt]
(axis cs:42,0.00565578131519695)
--(axis cs:42,0.00566901440480305);

\path [draw=gray119, line width=1.5pt]
(axis cs:46,0.00674707427642512)
--(axis cs:46,0.00687740933557488);

\path [draw=gray119, line width=1.5pt]
(axis cs:50,0.0082454789877193)
--(axis cs:50,0.0085488989442807);

\addplot [line width=1.5pt, chocolate2267451, dashed, mark=*, mark size=2.5, mark options={solid}]
table {%
10 0.000391783493877551
14 0.000927839746938776
18 0.00158100849387755
22 0.00213305639591837
26 0.00330323932244898
30 0.00438299699591837
34 0.00519606824489796
38 0.00620732203265306
42 0.00827288134693878
46 0.00975638982857143
50 0.0116961574081633
};
\addlegendentry{OSQP}
\addplot [line width=1.5pt, steelblue52138189, dashed, mark=*, mark size=2.5, mark options={solid}]
table {%
10 0.000152689248979592
14 0.000346945828571429
18 0.000767718897959184
22 0.00114121386530612
26 0.00172161326938776
30 0.00232782554693878
34 0.00297055062857143
38 0.00396053876734694
42 0.00485501457142857
46 0.0062998646122449
50 0.00773786895102041
};
\addlegendentry{ProxQP}
\addplot [line width=1.5pt, mediumpurple152142213, dashed, mark=*, mark size=2.5, mark options={solid}]
table {%
10 0.000120391836790819
14 0.000177594776650206
18 0.000220951249369192
22 0.000241695869267367
26 0.000298122841596356
30 0.000373062791216083
34 0.00042900179550812
38 0.000511097599519417
42 0.000589591771071986
46 0.000646835331192088
50 0.000737636702964367
};
\addlegendentry{ReLU-QP}
\addplot [line width=1.5pt, gray119, dashed, mark=*, mark size=2.5, mark options={solid}]
table {%
10 0.000235652304
14 0.0004204327
18 0.000670125874
22 0.001054915934
26 0.00165736904
30 0.00244941506
34 0.00350529079
38 0.00435558896
42 0.00566239786
46 0.006812241806
50 0.008397188966
};
\addlegendentry{OSQP-Direct}
\end{axis}

\end{tikzpicture}

%% file: sections/background_related_works.tex
\section{Background}\label{sec:background}

\subsection{Convex Quadratic Programs}
Convex quadratic programs are defined as follows \cite{nocedal_quadratic_2006}:
\begin{mini}|l|
    {y}{\frac{1}{2}y^T H y + g^T y}{}{}
    \addConstraint{c \leq Gy \leq d ,}
    \label{eq:qp_def}
\end{mini}
where $g \in \mathbb{R}^n $ and the symmetric positive-definite matrix $H \in \mathbb{R}^{n \times n}$ definite the cost function, $G \in \mathbb{R}^{m \times n}$, 
$c \in \mathbb{R}^m$ and $d \in \mathbb{R}^m$ 
define the constraints, and $n$ and $m$ are the number of decision variables and constraints, respectively. 

\subsection{Alternating Direction Method of Multipliers}\label{sec:admm}
The Alternating Direction Method of Multipliers (ADMM) is an efficient approach for solving \eqref{eq:qp_def} that has been widely deployed in robotics and machine learning in recent years \cite{stellato_osqp_2020}.
The ADMM algorithm introduces ``splitting variables'' $\bar{y}$ and $z$ to transform \eqref{eq:qp_def} into 

\begin{mini}|l|
    {y, \bar{y}, z}{\frac{1}{2}\bar{y}^TH\bar{y} + g^T\bar{y}}{}{}
    \addConstraint{G\bar{y}}{ = z}
    \addConstraint{y}{ = \bar{y}}
    \addConstraint{c}{ \leq z \leq d .}
    \label{eq:admm_qp_def}
\end{mini}

The augmented Lagrangian for \eqref{eq:admm_qp_def} is,
\begin{align}
\begin{split}
L_\rho (y, z, \lambda) = \frac{1}{2}y^THy + g^Ty + \frac{\sigma}{2}||\bar{y} - y||^2\\ + \mu^T(\bar{y} - y) + \frac{1}{2}(G\bar{y} - z)^T\rho(G\bar{y} - z) \\ + \lambda^T(G\bar{y} - z) + \mathbb{I}_{(c, d)}(z) ,
\label{eq:admm_al}
\end{split}
\end{align}
where $\sigma \in \mathbb{R}$ is a penalty weight, $\rho$ is a diagonal penalty matrix, $\mu$ and $\lambda$ are Lagrange multipliers, and $\mathbb{I}$ is the indicator function:

\begin{align}
    \mathbb{I}_{(c, d)}(z) = \begin{cases}
       0& c \leq z \leq d\\
       \infty& otherwise
    \end{cases} 
    \label{eq:indicator_func}
\end{align}
The diagonal elements in $\rho$ are used to weight progress towards satisfying each constraint. In our implementation the penalty on the equality constraints is a thousand times larger than the penalty on the inequality constraints to bias the algorithm towards converging to tighter tolerances on equality constraints.

ADMM alternates minimizing over $y$ and $z$ by performing the following steps \cite{stellato_osqp_2020}:
\begin{align}
\begin{split}
    \bar{y}^+ = &\ (H + \sigma I + G^T\rho G)^{-1}(-g + \sigma y - \mu  \\ & + G^T(\rho z - \lambda)) ,
\end{split}  \label{eq:Xbar_update} \\
\begin{split}
    y^+  = &\ \bar{y}^+ + \sigma^{-1}\mu ,
\end{split}  \label{eq:x_update} \\
    z^+ = &\ \Pi_{(c, d)}(G\bar{y}^+ + \rho^{-1}\lambda) , \label{eq:z_update} \\
    \mu^+ = &\ \mu + \sigma(\bar{y}^+ - y^+) , \label{eq:mu_update_gen} \\
    \lambda^+ = &\ \lambda + \rho(Gy^+ - z^+) , \label{eq:lambda_update_gen}
\end{align}
where $\Pi_{(c, d)}$ is a projection onto the box constraints $c \leq z \leq d$,
\eqref{eq:Xbar_update} is usually computed through a direct matrix factorization or using an indirect method like conjugate gradient, and steps \eqref{eq:x_update}-\eqref{eq:lambda_update_gen} are composed of simple addition, multiplication, and clamping operations. 
We note that, from the second iteration onwards, both equations \eqref{eq:mu_update_gen} and \eqref{eq:x_update} ensure that $\mu = 0$. Consequently, $\bar{y}$ and $y$ are effectively the same, making it possible to eliminate $y$ and $\mu$ and simplify the overall algorithm, leaving us with $\bar{y}$, $z$, and $\lambda$ as the essential variables.

\subsubsection{Convergence Criteria}
A natural choice of stopping criterion of the ADMM algorithm is the $\infty$-norm of the primal and dual residuals of \eqref{eq:admm_qp_def} \cite{boyd_distributed_2011}:
\begin{align}
    &\ ||Gy - z||_\infty \leq \epsilon_{prim}, \label{eq:primal}\\
    &\ ||Hy + g + G^T\lambda||_\infty \leq \epsilon_{dual}, \label{eq:dual}
\end{align}
where $\epsilon_{prim} \geq 0$ and $\epsilon_{dual} \geq 0$ are user-specified accuracy tolerances.

\subsubsection{Preconditioning} 
Numerical conditioning plays a key role in the convergence behavior of ADMM \cite{stellato_osqp_2020}.
Heuristic preconditioners \cite{giselsson_metric_2015, giselsson_linear_2017} are often used to reduce condition numbers and increase convergence rate. Variants of Ruiz equilibration \cite{ruiz_scaling_2001} are common in modern solvers \cite{stellato_osqp_2020, bambade_prox-qp_2022}.

\subsubsection{Adaptive Penalty}
The most important parameter in the ADMM algorithm is the parameter $\rho$ \cite{stellato_osqp_2020}.
Empirically, converge rates can be slow when a fixed penalty parameter is used \cite{boyd_distributed_2011}. A common heuristic that works well in practice is to adapt the penalty based on the ratio between the primal and dual residuals \cite{he_alternating_2000}. 
OSQP \cite{stellato_osqp_2020} implements a similar strategy that adjusts $\rho$ based on the relative magnitudes of the residuals:
\begin{align}
    \rho^+ = \rho\sqrt{\frac{||r_p||_\infty \max (||Hy||_\infty,||G^T\lambda||_\infty,||g||_\infty,10^{-4})}{||r_d||_\infty\max (||Gy||_\infty,||z||_\infty,10^{-4})}} ,\label{eq:rho_update}
\end{align}
where $r_p$ and $r_d$ are the primal dual residuals from \eqref{eq:primal} and \eqref{eq:dual}.
While adjusting the penalty term online is critical for fast convergence, each update requires a new matrix factorization, which is computationally expensive. Therefore, minimizing overall solution time requires a carefully balanced strategy. 

\subsubsection{Warm Starting}
When solving many similar problem instances sequentially, a standard warm-starting strategy for ADMM is to initialize the solver with $y$, $\lambda$, and $\rho$ from the previous solution and set $z = Gy$.

\subsection{Model-Predictive Control}
Linear MPC solves the receding-horizon optimal control problem:
\begin{mini}|l|
    {x_{1:N}, u_{0:N-1}}{\sum_{k=0}^{N-1}(\frac{1}{2}x_k^TQx_k + \frac{1}{2}u_k^TRu_k)+ \frac{1}{2}x_N^TQ_Nx_N}{}{}
    \addConstraint{x_{k + 1} = }{\;Ax_k + Bu_k \qquad \forall k \in {1, \dots, N - 1}}
    \label{eq:mpc_def}
\end{mini}
where $x_0$ is the current state of the system, $Q$ and $R$ are stage cost weights on the states and controls, respectively, $Q_N$ is the terminal state cost weight, $A$ is the discrete state transition matrix, $B$ is the discrete control Jacobian, and $N$ is the MPC horizon length. 
This optimization problem can be put into the standard form of \eqref{eq:qp_def} with the following cost and constraint matrices: 
\begin{gather*}
    y = \begin{bmatrix} u_0 \\ x_1 \\ u_1 \\ \vdots \\ x_N \end{bmatrix} ,\qquad
    H = \begin{bmatrix} 
        R & 0 & 0 & \dots & 0 \\
        0 & Q & 0 & \dots & 0 \\
        0 & 0 & R & \dots & 0 \\
        \vdots & \vdots & \vdots & \ddots & \vdots \\
        0 & 0 & 0 & \dots & Q_N \\
    \end{bmatrix} , \\
    G = \begin{bmatrix}
        B & -I & 0 & 0 & \dots & 0 \\
        0 & A & B & -I & \dots & 0 \\
        \vdots & \vdots & \vdots & \vdots & \ddots & \vdots \\
        0 & 0 & 0 & 0 & \dots & -I
    \end{bmatrix} , \quad c = d = \begin{bmatrix} -Ax_0 \\ 0 \\ \vdots \\ 0 \end{bmatrix} ,
\end{gather*}
where $G$, $c$, and $d$ encode the dynamics constraints, current state, and potentially additional state and control constraints. In this form, it is also straightforward to add additional constraints on the state and control variables. In the absence of additional constraints, \eqref{eq:mpc_def} is a Linear-Quadratic Regulator (LQR) problem and can be efficiently solved with a Riccati recursion \cite{scokaert_constrained_1998}.

We refer to \eqref{eq:mpc_def} as the ``direct'' formulation of the MPC problem, as the decision variables include both states and controls.
In practice, an MPC controller repeatedly re-solves this problem using the latest state information to produce a feedback policy. 

%% file: sections/technical_approach.tex
\section{The ReLU-QP Algorithm}
\label{sec:reluqp}

In this section, we derive a reformulation of the ADMM algorithm as a deep, weight-tied neural network with ReLU activations. This reformulation involves only GPU-friendly operations such as matrix-vector products and projection onto the positive orthant (clamping negative values to zero). 

\subsection{Modified ADMM Iterations}
We make two key modifications to the ADMM iteration in Section \ref{sec:admm}. First, we move the $\lambda$ update before the $y$ and $z$ updates:
\begin{align}
\begin{split}
    \lambda^+ = &\ \lambda + \rho(Gy - z) ,
\end{split} \label{eq:basic_lambda_update} \\
\begin{split} \label{eq:basic_y_update}
    y^+ = &\ (H + \sigma I + G^T\rho G)^{-1}(-g + \sigma y  \\ & + G^T(\rho z - \lambda^+)) ,
\end{split}\\
\begin{split} \label{eq:clamp}
    z^+ = &\ \Pi_{(c, d)}(Gy^+ + \rho^{-1}\lambda^+) .
\end{split}
\end{align}
We then combine \eqref{eq:basic_lambda_update} and \eqref{eq:basic_y_update} into a single matrix equation, followed by the projection \eqref{eq:clamp}, where $\tilde{c}$ and $\tilde{d}$ are $(-\infty, +\infty)$ for $y$ and $\lambda$:
\begin{align}
    \begin{bmatrix} y^+ \\ z^+ \\ \lambda^+ \end{bmatrix} = \Pi_{(\tilde{c},\tilde{d})} \left( W\begin{bmatrix} y \\ z \\ \lambda\end{bmatrix} + b \right) ,
    \label{eq:gpu_update}
\end{align}
where
\begin{gather*}
    W = \\\begin{bmatrix}
        D(\sigma I - G^T\rho G)       & 2DG^T\rho       & -DG^T \\
        GD(\sigma I - G^T\rho G) + G  & 2GDG^T\rho - I  & -GDG^T + \rho^{-1} \\
        \rho G                     & -\rho           & I
    \end{bmatrix} ,\\
    D = (H + \sigma I + G^T\rho G)^{-1} , \qquad
    b = \begin{bmatrix}
        -D \\ -GD \\ 0
    \end{bmatrix}g .
\end{gather*}

Note that, in spite of some additional computation introduced by reordering the ADMM update steps, the matrix-vector product and projection operations in \eqref{eq:gpu_update} can be easily parallelized row-wise, and can therefore be efficiently computed on a GPU.

\subsection{Algorithm Summary}
The ReLU-QP algorithm includes both offline and online stages: The offline stage involves pre-computing a set of matrices $W$ in \eqref{eq:gpu_update} for a predefined set of penalty parameters. In the online stage, we initialize the primal and dual variables and then perform the modified ADMM iterations \eqref{eq:gpu_update} until convergence criteria \eqref{eq:primal} and \eqref{eq:dual} are satisfied, as summarized in Algorithm \ref{alg:gpu_admm}.

\begin{algorithm}
\caption{ReLU-QP}\label{alg:gpu_admm}
\begin{algorithmic}
    \State \textbf{Given:} $\epsilon$, $\rho_i$, check\_interval, max\_iters
  \State \textbf{Offline:}
    \State Scale problem
    \State Compute KKT inverse, $W$ and $b$ for each $\rho \in \rho s$
  \State \textbf{Online:}
  \State Update $b$, $\tilde c$, and $\tilde d$ with initial conditions
  \State $v = \left[ y; \; z; \; \lambda \right]$
  \For{i = 1:max\_iters}
    \State $v \leftarrow \text{clamp}(W v + b, \tilde{c}, \tilde{d})$
    \If{mod(i, check\_interval) == 0}
    \State W, b $\leftarrow$ rho\_update($v$)
        \If{residuals($v$) $< \epsilon $}
            break
    \EndIf\EndIf\EndFor
    \State return y
\end{algorithmic}
\end{algorithm}

\subsection{Implementation details}
\subsubsection{Penalty Scaling Heuristics}
By default, we define a list of penalty weights $\rho_i$ sampled logarithmically from $10^{-3}$ to $10^{3}$ and set the initial value to $\rho = 0.1$.

During the solve, we compute a nominal penalty parameter from \eqref{eq:rho_update} and use the closest value in the predefined list by choosing the corresponding weight $W$ and bias $b$.

\subsubsection{Preconditioning} \label{sec:preconditioning}
We use a modified Ruiz equilibration heuristic \cite{ruiz_scaling_2001} for ill-conditioned problems. Similar to OSQP, \cite{stellato_osqp_2020}, we scale both $D$ and the cost to improve the convergence rate.

\subsubsection{Convergence Checking}
The conventional wisdom that ADMM iterations are computationally inexpensive relative to computing convergence criteria \eqref{eq:primal} and \eqref{eq:dual} still holds for our solver. We check the convergence criteria every $25$ iteration by default.

\subsubsection{Software Implementation}
We implement the ReLU-QP solver in three popular machine-learning libraries: Flux.jl \cite{innes_flux_2018}, PyTorch \cite{paszke_pytorch_2019}, and JAX \cite{noauthor_jax_2023}. All three implementations demonstrate similar performance, especially on large problems where the solve time is dominated by low-level GPU operations on large matrices.  Open-source implementations of ReLU-QP can be found at: \url{https://github.com/RoboticExplorationLab/ReLUQP.jl}.

%% file: sections/MPC.tex
\section{Dense Model-Predictive Control}
\label{sec:mpc}
We solve a condensed form of the MPC problem \eqref{eq:mpc_def} to take advantage of existing GPU-accelerated dense matrix operations from standard machine-learning libraries. A common technique for condensing MPC problems is to eliminate the states as variables and only optimize over the controls given an initial condition $x_0$. This idea is similar to single shooting methods where the solver only optimizes over dynamically feasible trajectories, resulting in fewer decision variables and reduced sparsity structures. However, for open-loop unstable systems, condensing the MPC problem in this way also introduces severe numerical ill-conditioning  \cite{tedrake2019}.  

We find that the modified Ruiz equilibration is often insufficient for addressing ill-conditioning in these problems. Instead, we use an infinite-horizon LQR feedback controller as an effective preconditioner, expressing the controls as $u_k = -Kx_k + \Delta u_k$, where $K$ is the LQR gain matrix and $\Delta u_k$ are the new decision variables in the MPC problem. This modification has been shown to eliminate the ill-conditioning associated with condensed MPC problems and improve solver convergence \cite{rossiter_numerically_1998, mcinerney_horizon-independent_2023}.

In summary, we define the following transformations:
:
\begin{align*}
    y = &\ S\tilde{y} + Mx_0 , \qquad \bar{A} = A - BK ,\\
    \tilde{y} = &\ \begin{bmatrix} \Delta u_1 \\ \Delta  u_2 \\ \vdots \\ \Delta u_N \end{bmatrix} , \qquad 
    M =  \begin{bmatrix}
        -K \\ \bar{A} \\ 
        \vdots \\ -K\bar{A}^{N - 1} \\ \bar{A}^N
    \end{bmatrix} ,\\
    S = &\ \begin{bmatrix}
        I                   & 0                   & \hdots & 0\\
        B                   & 0                   & \hdots & 0\\
        \vdots              & \vdots              & \ddots & \\
        -K\bar{A}^{N - 2}B & -K\bar{A}^{N - 3}B & \hdots & I\\
        \bar{A}^{N - 1}B   & \bar{A}^{N - 2}B   & \hdots & B
    \end{bmatrix} . \\
\end{align*}
The Hessian, gradient, and constraint matrices of the ``direct'' form are then modified as follows: 
\begin{align*}
    \bar{H} = S^THS, \qquad \bar{g} = S^THMx_0, \qquad \bar{G} = GS, \\
    \qquad \bar{c} = c - GMx_0, \qquad \bar{d} = d - GMx_0.
\end{align*}
The rows of $G$ corresponding to the dynamics constraints are eliminated in this formulation as they are included in the new cost Hessian.

%% file: sections/experiments_and_results.tex
\section{Simulation Experiments} \label{sec:experiments}

In this section, we present the results of benchmarking experiments demonstrating ReLU-QP on both random, dense QPs and simulated closed-loop MPC problems, including balancing the Atlas humanoid on one foot and reference tracking on a quadruped robot equipped with a six-degree-of-freedom arm.

All experiments were performed on a desktop equipped with an Intel i7-12700K CPU with 64GB of memory and an NVIDIA GeForce RTX 3080 GPU with 10GB of VRAM. For OSQP \cite{stellato_osqp_2020} and (dense) ProxQP \cite{bambade_prox-qp_2022}, we report internal solver timings and use the same convergence tolerances across all solvers to ensure fair comparisons. Unless otherwise specified, we solve each problem three times and average the solution times. All benchmarks are run in Julia \cite{koolen_julia_2019} with the Flux.jl \cite{innes_flux_2018} implementation of ReLU-QP. 

\subsection{Random Dense QPs} \label{random_qp}
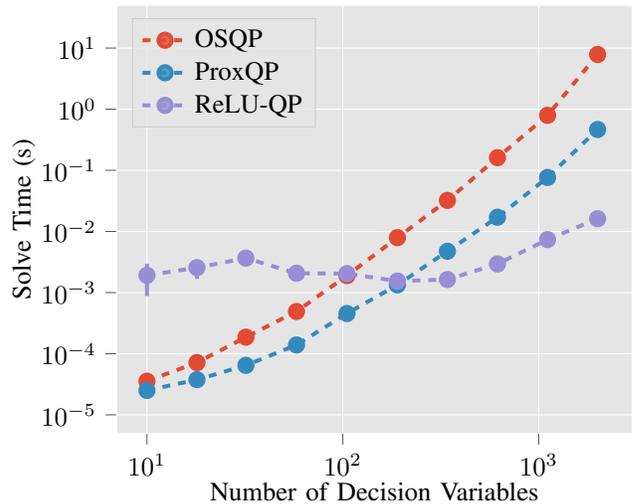
\begin{figure}
    \centering
    \input{figures/random_mixed_qp_init_high_tol}
    \caption{Solve times for random dense QPs with both equality and inequality constraints demonstrating that ReLU-QP is up to $30$ times faster than OSQP and ProxQP on large problems. 
    }
    \label{fig:rand_init_solve}
\end{figure}

We first benchmark on random dense QPs with both equality and inequality constraints. We generated problems for ten different decision-variable dimensions ($n$) logarithmically spaced from $10$ to $2000$. The number of equality and inequality constraints in each problem are each $n/4$. For each problem size, we randomly generate ten dense QPs and solve each problem until the residuals are below $10^{-6}$. Note that, with the exception of tolerance settings, default parameters are used for all solvers. Timing results are shown in Fig. \ref{fig:rand_init_solve}. Compared to OSQP \cite{stellato_osqp_2020} and ProxQP \cite{bambade_prox-qp_2022}, our method provides competitive solve times on medium-sized problems ($n = 100$) and is up to $30$ times faster on large ($n=2000$) problems.

\subsection{Random Linear MPC Problems}

We randomly generate controllable linear systems with control dimensions ranging from $10$ to $50$, where the state dimensions are always chosen to be triple the control dimensions, by sampling the $A$ and $B$ matrices in \eqref{eq:mpc_def}. We then use MPC to drive each system to the origin under control limits from a random initial state, making sure the initial states were large enough to activate the constraints for a significant portion of the trajectory. For OSQP and ReLU-QP, we perform one solver iteration per MPC step. For ProxQP, we solve to $10^{-2}$ tolerance, which normally corresponds to $1$ outer iteration. All solvers are warm started with the primal and dual solutions from the previous MPC step. Solve times are shown in Fig. \ref{fig:random_mpc}, where ReLU-QP is an order of magnitude faster on high-dimensional problems. 

Note that all experiments in Fig. \ref{fig:random_mpc} solve the LQR-preconditioned dense MPC formulation, except \emph{OSQP-Direct}, which solves the ``direct'' MPC formulation \eqref{eq:mpc_def} using a sparse linear system solver (QDLDL) \cite{stellato_osqp_2020} that efficiently exploits matrix sparsity. 

\subsection{Atlas Balance}

We use MPC to balance a full-body model of the Atlas humanoid robot balancing on one-foot subject to initial velocity disturbances and control limits. The model has $58$ states and $29$ control inputs, and we linearize about a nominal reference pose.
We benchmark our solver against OSQP and ProxQP using the preconditioned condensed formulation, with MPC horizons of $0.3$s, $0.4$s, and $0.5$s. We warm start each solver by solving an initial problem instance to $10^{-6}$ tolerances. After the initial solve, we run OSQP and ReLU-QP for $2$ iterations per MPC step and ProxQP to $10^{-4}$ tolerances -- both of which were empirically found to be the coarsest and fastest settings that could successfully stabilize the robot.

\begin{figure}
    \centering
    \begin{overpic}[trim=10cm 30cm 5cm 3.0cm, clip, height=3.75cm]{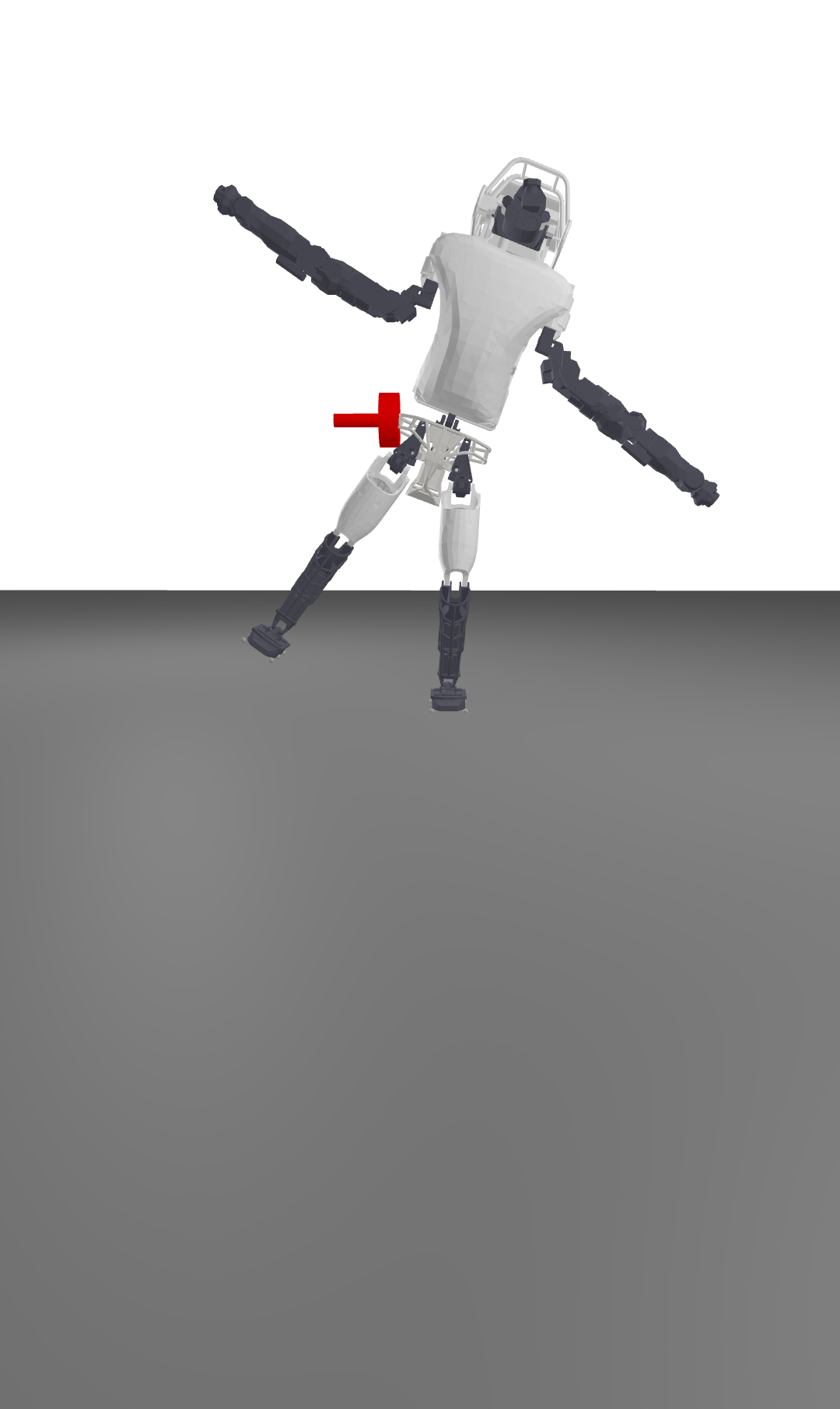}
    \put(5,100){$t=0.2$s}
  \end{overpic}
    \hfill
    \begin{overpic}[trim=10cm 30cm 5cm 3.0cm, clip, height=3.75cm]{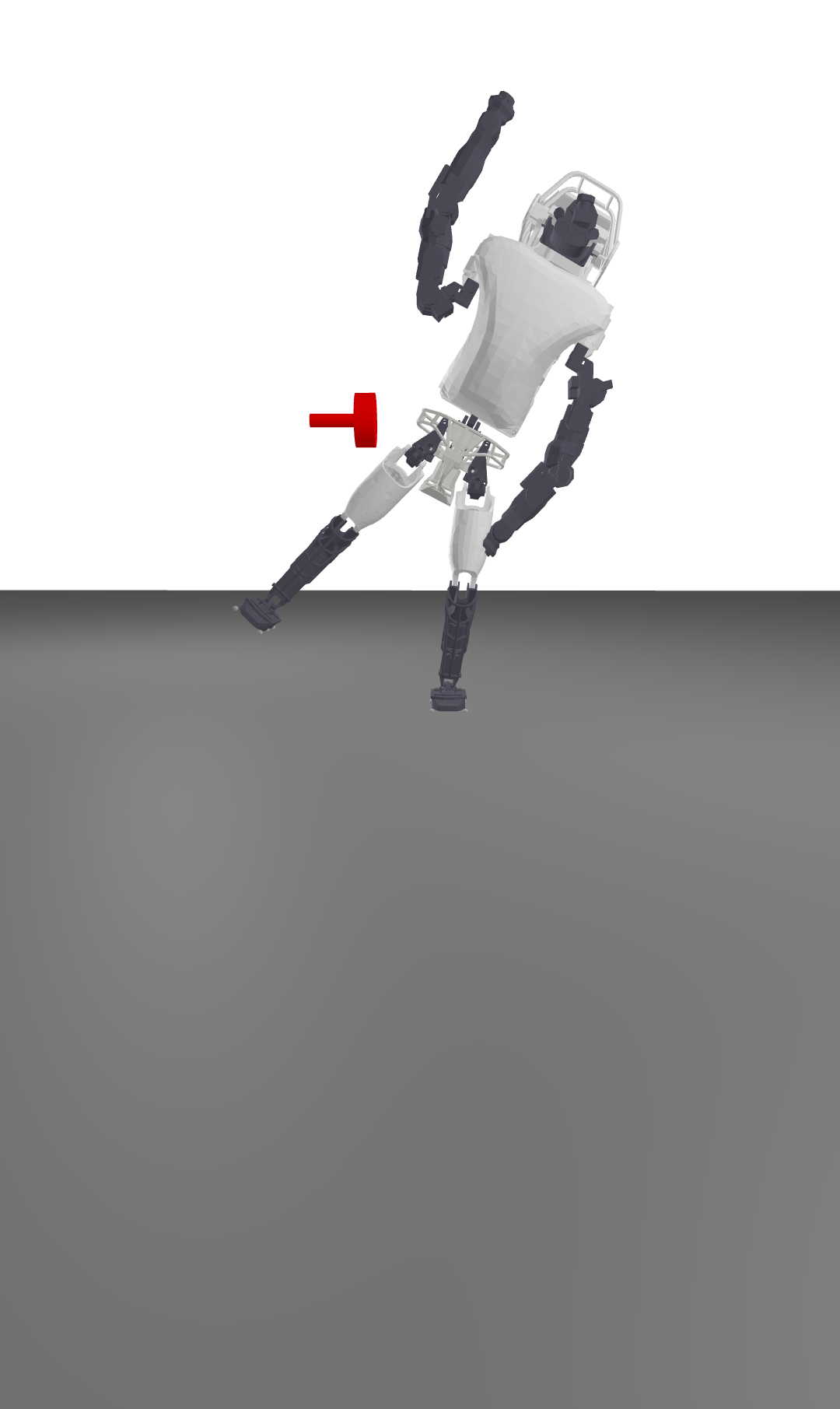}
    \put(5,100){$t=1.1$s}
  \end{overpic}
    \hfill
    \begin{overpic}[trim=10cm 30cm 5cm 3.0cm, clip, height=3.75cm]{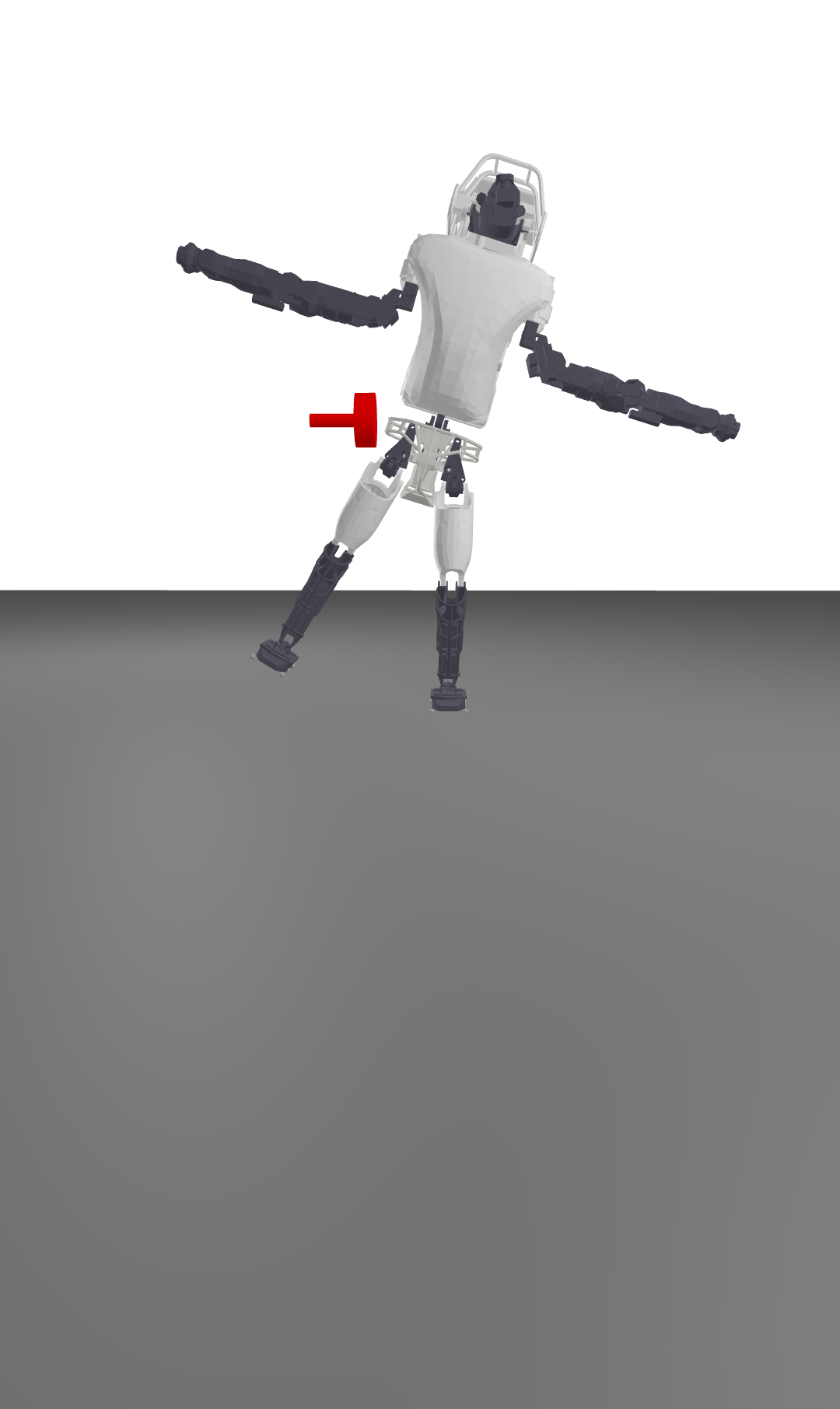}
    \put(5,100){$t=5.0$s}
  \end{overpic}
    \caption{An Atlas humanoid robot balancing on one foot subject to control limits and disturbances. ReLU-QP successfully solves the task at up to $2600$ Hz.}
    \label{fig:atlas_balance}
\end{figure}

Timing results for each horizon length are shown in Table \ref{table:atlas}. ReLU-QP demonstrates speed improvements of up to $20$ times and is able to successfully stabilize the non-linear Atlas model at up to $2600$ Hz, introducing the possibility of using whole-body MPC on high-dimensional robots without lower-level joint-space or impedance controllers. 

\begin{table}[t]
\centering
\caption{Average MPC solve frequencies for a whole-body Atlas stabilizing task. ReLU-QP demonstrates $10$-$20$ times speed ups.}
    \begin{tabular}{c c c c c}
    \toprule
    \textbf{MPC Horizon} & \textbf{OSQP} & \textbf{ProxQP} & \textbf{ReLU-QP} & \textbf{speed up}\\
    \midrule
    $0.3$ s  & $206$ Hz & $192$ Hz & $\textbf{2599}$ Hz & \textbf{10x}\\ 
    $0.4$ s  & $104$ Hz & $153$ Hz & $\textbf{1925}$ Hz & \textbf{13x}\\
    $0.5$ s  & $65$ Hz & $66$ Hz & $\textbf{1348}$ Hz & \textbf{20x}\\
    \bottomrule
    \end{tabular}

\label{table:atlas}
\end{table}

\subsection{Reference Tracking on a Quadruped with an Arm}
We used MPC to control a quadruped pick-up motion with a 6 DoF arm. The model has $52$ states, $20$ control inputs, and $12$ foot-contact force variables, and was linearized around a nominal standing pose. We added pyramidal friction cone, normal force, and pinned-foot constraints as well as torque limits, totaling $52$ constraints at each time step point. We benchmarked ReLU-QP against ProxQP and OSQP, with ReLU-QP and ProxQP solving the preconditioned condensed formulation and OSQP solving the sparse direct formulation. Each solver was warm-started with ReLU-QP and OSQP running for $15$ iterations per MPC step, which were empirically found to be the coarsest and fastest settings that could achieve the task. However, even when solving to a tolerance of $10^{-6}$ we were unable to stabilize the system using ProxQP. The results of the benchmarks are shown in Table \ref{table:quadruped}. ReLU-QP was able to run over two times faster than OSQP with a frequency of 834 Hz, compared to 325 Hz for OSQP.
\begin{figure}
    \begin{overpic}[trim=6cm 12.0cm 4cm 2cm, clip, height=3.75cm]{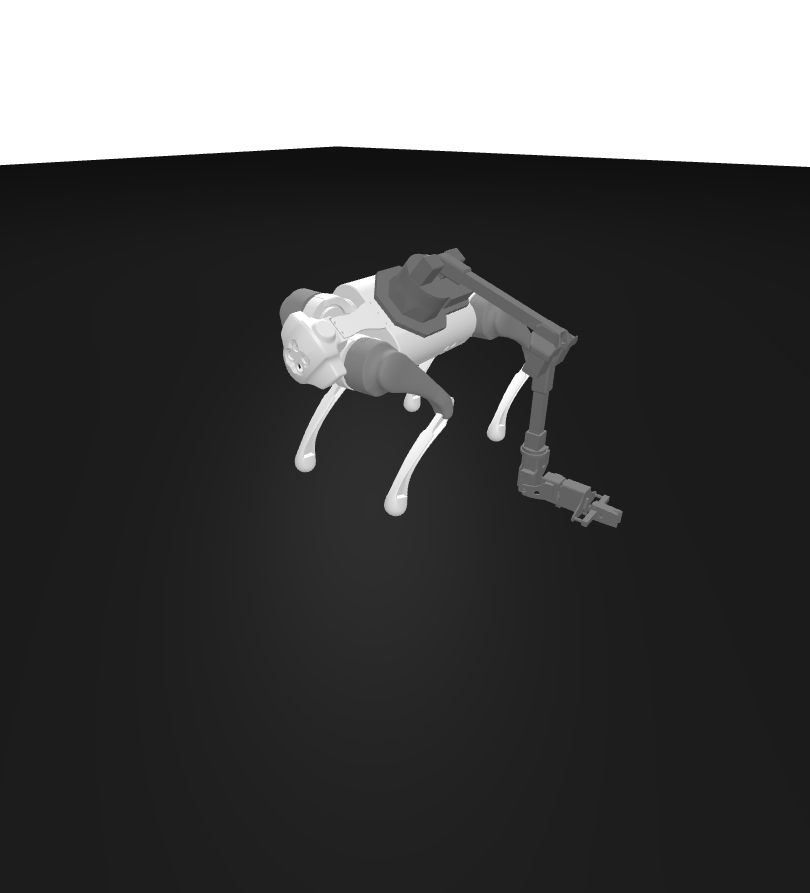}
    \put(70,95){$t=0.0$s}
  \end{overpic}
  \hfill
    \begin{overpic}[trim=6cm 12.0cm 4cm 2cm, clip, height=3.75cm]{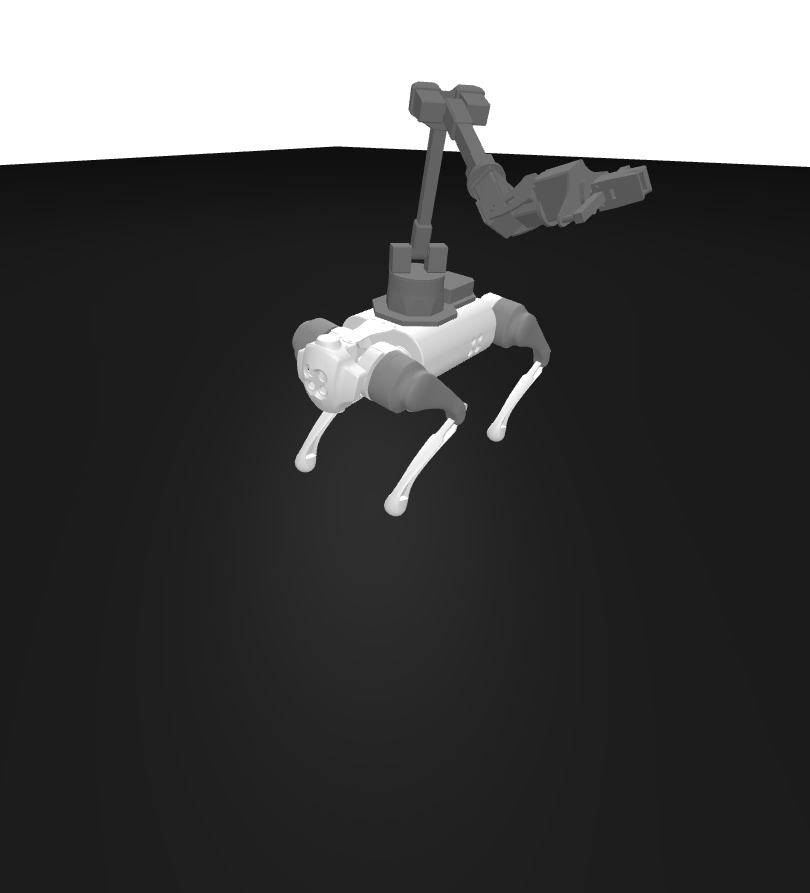}
    \put(70,95){$t=0.2$s}
  \end{overpic}
    \begin{overpic}[trim=6cm 12.0cm 4cm 2cm, clip, height=3.75cm]{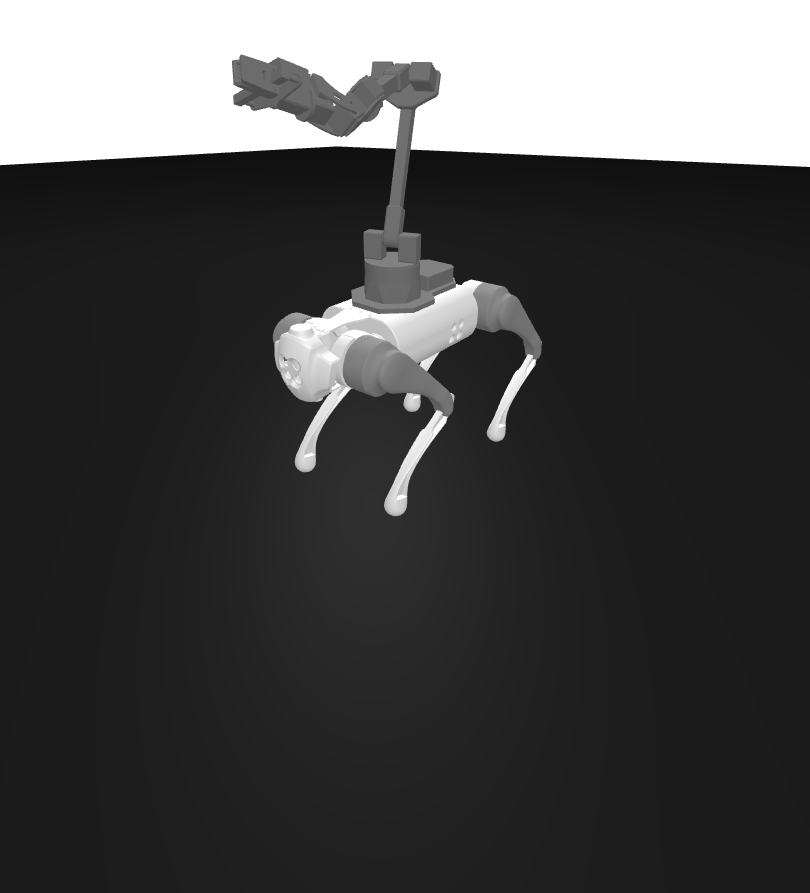}
    \put(70,95){$t=0.4$s}
  \end{overpic}
  \hfill
    \begin{overpic}[trim=6cm 12.0cm 4cm 2cm, clip, height=3.75cm]{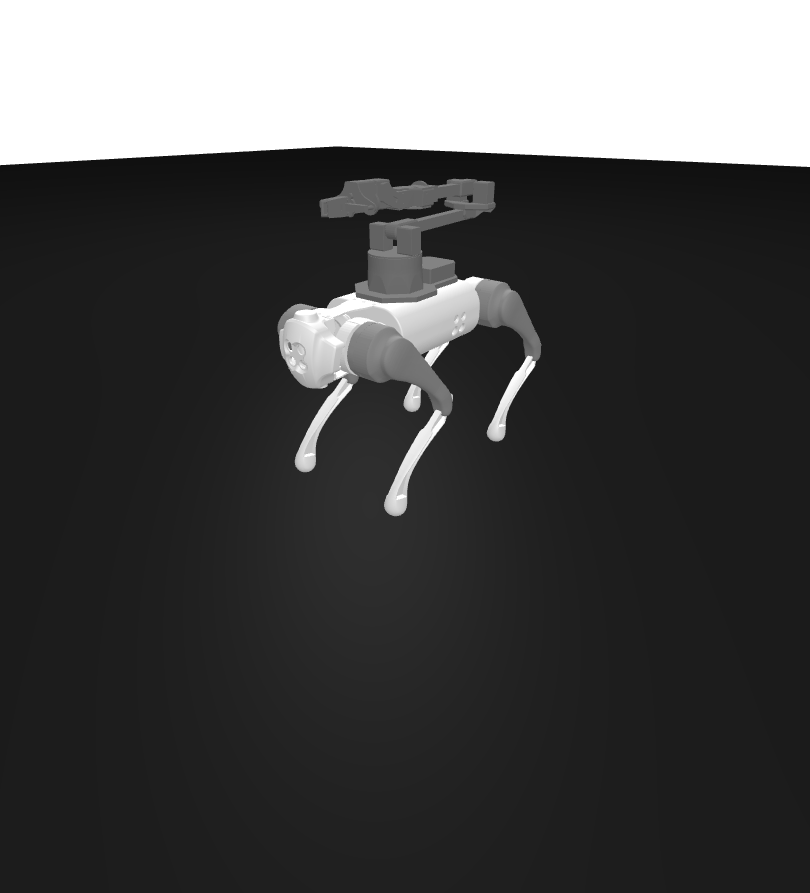}
    \put(70,95){$t=2.2$s}
  \end{overpic}

    \caption{A Go1 robot equipped with a $6$ DoF arm performing a pick-up motion using linear MPC. ReLU-QP solves the problem at 780 Hz.}
    \label{fig:enter-label}
\end{figure}

\begin{table}[t]
\centering
\caption{Average MPC solve frequencies for a whole-body quadruped model with an arm picking up an object from the ground. ReLU-QP demonstrates a $2 \times$ speed up over OSQP.}
    \begin{tabular}{c c c c c}
    \toprule
    \textbf{MPC Horizon} & \textbf{OSQP} & \textbf{ProxQP} & \textbf{ReLU-QP} & \textbf{speed up}\\
    \midrule
    $0.3$ s  & $325$ Hz & -- Hz & $\textbf{834}$ Hz & \textbf{2.6x}\\ 
    \bottomrule
    \end{tabular}

\label{table:quadruped}
\end{table}

%% file: figures/random_mixed_qp_init_high_tol.tex
\begin{tikzpicture}

\definecolor{chocolate2267451}{RGB}{226,74,51}
\definecolor{dimgray85}{RGB}{85,85,85}
\definecolor{gainsboro229}{RGB}{229,229,229}
\definecolor{lightgray204}{RGB}{204,204,204}
\definecolor{mediumpurple152142213}{RGB}{152,142,213}
\definecolor{steelblue52138189}{RGB}{52,138,189}

\begin{axis}[
axis background/.style={fill=gainsboro229},
axis line style={white},
legend cell align={left},
legend style={
fill opacity=0.8,
draw opacity=1,
text opacity=1,
at={(0.03,0.97)},
anchor=north west,
draw=lightgray204,
fill=gainsboro229
},
log basis x={10},
log basis y={10},
tick align=outside,
tick pos=left,
x grid style={white},
xlabel={Number of Decision Variables},
xmajorgrids,
xmin=7, xmax=3000,
xmode=log,
xtick style={color=dimgray85},
xtick = {1e1, 1e2, 1e3},
y grid style={white},
ylabel={Solve Time (s)},
ymajorgrids,
ymin=5e-06, ymax=50,
ymode=log,
ytick style={color=dimgray85},
ytick={1e-5, 1e-4, 1e-3, 1e-2, 1e-1, 1e0, 1e1}
]
\path [draw=chocolate2267451, line width=1.5pt]
(axis cs:10,2.95368207377363e-05)
--(axis cs:10,4.1442712595597e-05);

\path [draw=chocolate2267451, line width=1.5pt]
(axis cs:18,6.08896919390117e-05)
--(axis cs:18,8.25044413943216e-05);

\path [draw=chocolate2267451, line width=1.5pt]
(axis cs:32,0.000164429077799008)
--(axis cs:32,0.000208774822200992);

\path [draw=chocolate2267451, line width=1.5pt]
(axis cs:58,0.000468242463932316)
--(axis cs:58,0.000514046769401017);

\path [draw=chocolate2267451, line width=1.5pt]
(axis cs:105,0.00179163721986487)
--(axis cs:105,0.0019980847468018);

\path [draw=chocolate2267451, line width=1.5pt]
(axis cs:190,0.00716347308885136)
--(axis cs:190,0.00866599344448197);

\path [draw=chocolate2267451, line width=1.5pt]
(axis cs:342,0.0281339820988203)
--(axis cs:342,0.0362795827678464);

\path [draw=chocolate2267451, line width=1.5pt]
(axis cs:616,0.130334625747094)
--(axis cs:616,0.191393541019573);

\path [draw=chocolate2267451, line width=1.5pt]
(axis cs:1110,0.735483241313331)
--(axis cs:1110,0.850208162286668);

\path [draw=chocolate2267451, line width=1.5pt]
(axis cs:2000,7.01919920745291)
--(axis cs:2000,8.68189232651375);

\path [draw=steelblue52138189, line width=1.5pt]
(axis cs:10,1.82938984833378e-05)
--(axis cs:10,3.18915681833289e-05);

\path [draw=steelblue52138189, line width=1.5pt]
(axis cs:18,3.22826296515655e-05)
--(axis cs:18,4.30170370151011e-05);

\path [draw=steelblue52138189, line width=1.5pt]
(axis cs:32,5.74504993892719e-05)
--(axis cs:32,7.16702672773948e-05);

\path [draw=steelblue52138189, line width=1.5pt]
(axis cs:58,0.000129513325232859)
--(axis cs:58,0.000148616074767141);

\path [draw=steelblue52138189, line width=1.5pt]
(axis cs:105,0.000422361900574807)
--(axis cs:105,0.00048727686609186);

\path [draw=steelblue52138189, line width=1.5pt]
(axis cs:190,0.00124258593197875)
--(axis cs:190,0.00142764493468792);

\path [draw=steelblue52138189, line width=1.5pt]
(axis cs:342,0.00459941385020515)
--(axis cs:342,0.00488704864979484);

\path [draw=steelblue52138189, line width=1.5pt]
(axis cs:616,0.0166830193112442)
--(axis cs:616,0.0174872755554224);

\path [draw=steelblue52138189, line width=1.5pt]
(axis cs:1110,0.0736998583090712)
--(axis cs:1110,0.0792153124575955);

\path [draw=steelblue52138189, line width=1.5pt]
(axis cs:2000,0.43544473413367)
--(axis cs:2000,0.495047536699663);

\path [draw=mediumpurple152142213, line width=1.5pt]
(axis cs:10,0.000879255790824081)
--(axis cs:10,0.00295368822147145);

\path [draw=mediumpurple152142213, line width=1.5pt]
(axis cs:18,0.00168086158720288)
--(axis cs:18,0.00343522906592177);

\path [draw=mediumpurple152142213, line width=1.5pt]
(axis cs:32,0.00299038050022516)
--(axis cs:32,0.00433945737507004);

\path [draw=mediumpurple152142213, line width=1.5pt]
(axis cs:58,0.0016213345812231)
--(axis cs:58,0.0025236937112791);

\path [draw=mediumpurple152142213, line width=1.5pt]
(axis cs:105,0.00158702840486545)
--(axis cs:105,0.00249136092950082);

\path [draw=mediumpurple152142213, line width=1.5pt]
(axis cs:190,0.00130784755248484)
--(axis cs:190,0.00177641805289677);

\path [draw=mediumpurple152142213, line width=1.5pt]
(axis cs:342,0.00156733559391376)
--(axis cs:342,0.00170452574675251);

\path [draw=mediumpurple152142213, line width=1.5pt]
(axis cs:616,0.00236991892651787)
--(axis cs:616,0.00351208108520391);

\path [draw=mediumpurple152142213, line width=1.5pt]
(axis cs:1110,0.00730064442159715)
--(axis cs:1110,0.00741455984673783);

\path [draw=mediumpurple152142213, line width=1.5pt]
(axis cs:2000,0.0160606050738763)
--(axis cs:2000,0.0162586164082508);

\addplot [line width=1.5pt, chocolate2267451, dashed, mark=*, mark size=2.5, line width=1.5pt, mark options={solid}]
table {%
10 3.54897666666667e-05
18 7.16970666666667e-05
32 0.00018660195
58 0.000491144616666667
105 0.00189486098333333
190 0.00791473326666667
342 0.0322067824333333
616 0.160864083383333
1110 0.7928457018
2000 7.85054576698333
};
\addlegendentry{OSQP}
\addplot [line width=1.5pt, steelblue52138189, dashed, mark=*, mark size=2.5, line width=1.5pt, mark options={solid}]
table {%
10 2.50927333333333e-05
18 3.76498333333333e-05
32 6.45603833333333e-05
58 0.0001390647
105 0.000454819383333333
190 0.00133511543333333
342 0.00474323125
616 0.0170851474333333
1110 0.0764575853833333
2000 0.465246135416667
};
\addlegendentry{ProxQP}
\addplot [mediumpurple152142213, dashed, mark=*, mark size=2.5,line width=1.5pt, mark options={solid}]
table {%
10 0.00191647200614777
18 0.00255804532656233
32 0.0036649189376476
58 0.0020725141462511
105 0.00203919466718313
190 0.00154213280269081
342 0.00163593067033313
616 0.00294100000586089
1110 0.00735760213416749
2000 0.0161596107410636
};
\addlegendentry{ReLU-QP}
\end{axis}

\end{tikzpicture}

%% file: sections/discussion_and_conclusions.tex
\section{Discussion and Conclusions}
\label{sec:conclusions}
In this paper, we present ReLU-QP, a GPU-accelerated quadratic programming solver capable of solving medium-to-large problem instances at real-time rates. Our solver excels at quickly solving large, dense QPs to coarse tolerances, and the LQR-preconditioned condensed MPC formulation enables high-performance model-predictive control on complex, high-dimensional systems.

\subsection{Limitations}
The primary limitation of our method is the need to pre-compute weight matrices offline, which limits our ability to perform online updates or solve nonlinear problems with sequential quadratic programming. Additionally, computing weight matrices for a large number of penalty parameters can consume an excessive amount of GPU VRAM.

We also note that the speed-ups achieved on the quadruped example were much less than on Atlas, which we attribute to the very large number of constraints in the problem and ReLU-QP's current inability to leverage matrix sparsity. OSQP, in contrast, uses a sparse linear system solver that can efficiently exploit the sparsity in these constraint equations.

\subsection{Future Work}
Several directions remain for future work.
First, we plan to extend ReLU-QP to better exploit problem sparsity.
Second, we will investigate indirect linear-system solvers like conjugate gradient to avoid pre-computing inverses offline and enable online updates and nonlinear SQP methods. 
Finally, we plan to investigate the possibility of initializing neural network control policies using ReLU-QP before performing additional policy tuning using reinforcement-learning techniques.